%% file: eccv_paper.tex
\documentclass[runningheads]{llncs}

\usepackage[table, dvipsnames]{xcolor}
\usepackage{eccv}

\usepackage{eccvabbrv}

\usepackage{graphicx}
\usepackage{booktabs}
\usepackage{comment}
\usepackage{siunitx}
\usepackage{colortbl}
\usepackage{booktabs}
\usepackage{graphicx}  
\usepackage{colortbl}
\usepackage{lipsum} 
\include{preamble}

\usepackage{wrapfig}

\definecolor{cvprblue}{rgb}{0.21,0.49,0.74}
\definecolor{lblue}{rgb}{0.9,0.95,1}
\definecolor{lpurple}{rgb}{0.35,0.25,0.55}
\definecolor{lgreen}{rgb}{0.95,1,0.95}
\definecolor{sblue}{rgb}{0,0.45,1}
\definecolor{gold}{RGB}{248, 214, 99}
\definecolor{bronze}{RGB}{231, 188, 133}
\definecolor{silver}{RGB}{198, 210, 226}

\usepackage[accsupp]{axessibility}  

\usepackage[pagebackref,breaklinks,colorlinks,citecolor=eccvblue]{hyperref}

\usepackage{orcidlink}

\begin{document}

\title{AIM 2024 Challenge on UHD Blind Photo Quality Assessment}

\titlerunning{AIM 2024 Blind UHD-IQA Challenge}

\author{
Vlad Hosu\inst{3}$^\dagger$\orcidlink{0000-0001-7070-5688} \and
Marcos V. Conde\inst{1,2}$^{\dagger \ddagger}$\orcidlink{0000-0002-5823-4964} \and
Lorenzo Agnolucci\inst{3,4}$^\dagger$\orcidlink{0000-0002-9558-1287} 
\and
Nabajeet Barman\inst{2}$^\dagger$\orcidlink{0000-0003-2587-7370}  \and
Saman Zadtootaghaj\inst{2}$^\dagger$\orcidlink{0000-0002-6028-8507} \and
Radu Timofte\inst{1}$^\dagger$\orcidlink{0000-0002-1478-0402}
\and \\
Wei Sun \and Weixia Zhang \and Yuqin Cao \and Linhan Cao \and Jun Jia \and Zijian Chen \and Zicheng Zhang \and Xiongkuo Min \and Guangtao Zhai \and
Songbai Tan \and Lixin Zhang \and Guanghui Yue \and
Daekyu Kwon \and Dongyoung Kim \and Seon Joo Kim \and
Yunchen Zhang \and Xiangkai Xu \and Hong Gao \and Yiming Bao \and Ji Shi \and Xiugang Dong \and Xiangsheng Zhou \and Yaofeng Tu \and
Zewen Chen \and
Shunhan Xu \and
Haochen Guo \and
Yun Zeng \and
Shuai Liu \and
Jian Guo \and
Juan Wang \and
Bing Li \and
Dehua Liu \and
Hesong Liu \and
Grigory Malivenko \and
Asile Gerek \and
Xingyuan Ma \and Cheng Li \and 
Joonhee Lee \and
Junseo Bang \and
Se Young Chun
}

\authorrunning{Hosu, Conde, Agnolucci, Barman, Zadtootaghaj, Timofte et al.}

\institute{
Computer Vision Lab, CAIDAS \& IFI, University of Würzburg \and
Visual Computing Group, FTG, Sony PlayStation \and 
Sony AI \and
University of Florence \\
$^\dagger$ Challenge Organizers, $^\ddagger$ Corresponding Author
\url{https://database.mmsp-kn.de/uhd-iqa-benchmark-database.html}
}

\maketitle


\vspace{-2.5mm}

\begin{figure}
    \centering
    \includegraphics[width=0.72\linewidth]{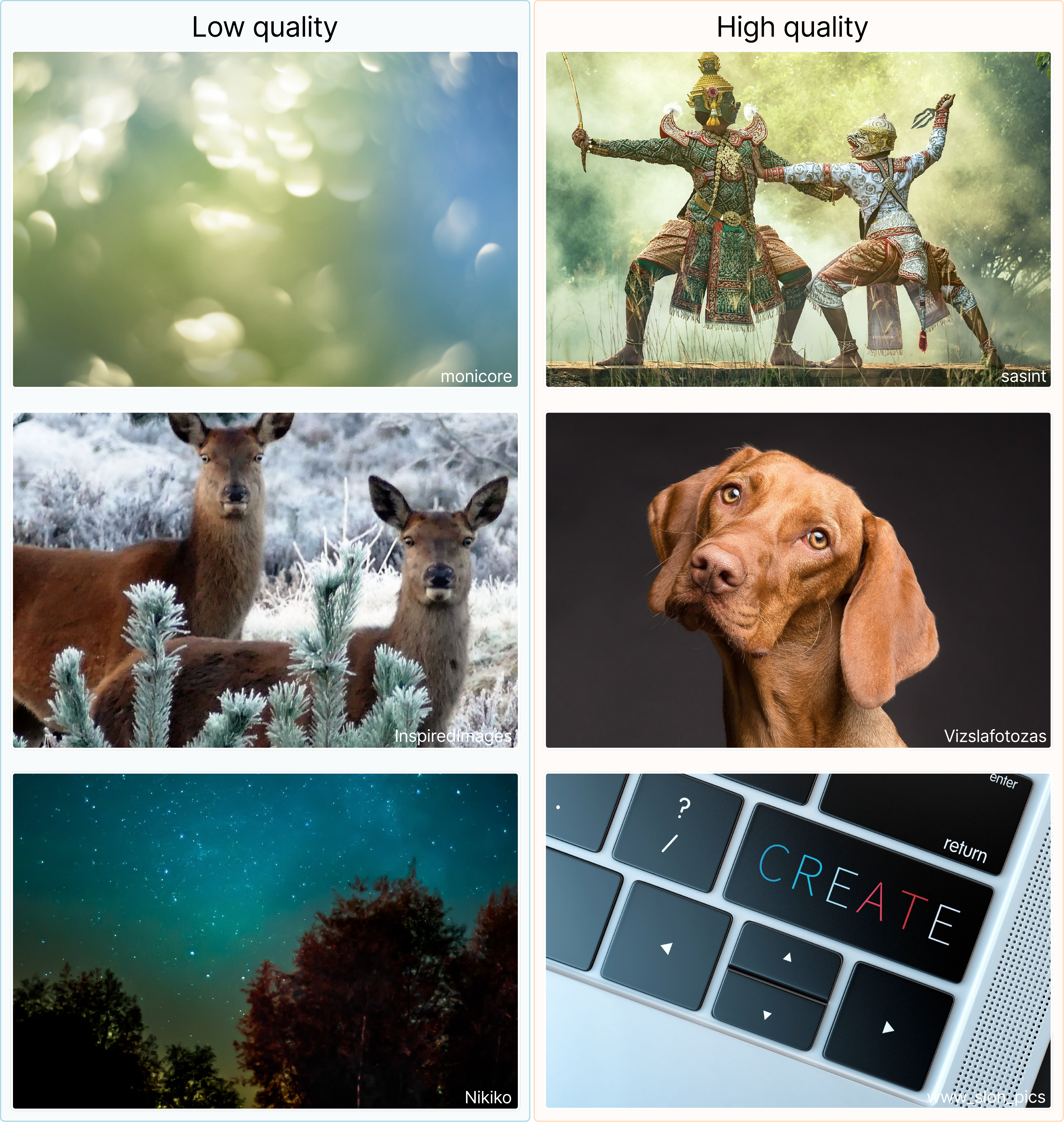}
    \caption{Example images from the UHD-IQA dataset \cite{hosu2024uhd}. They have been cropped to 64\% of their original size to enhance detail visibility. The author's name from \url{Pixabay.com} is shown at the bottom right of each image.}
    \label{fig:teaser}
    \vspace{-7.5mm}
\end{figure}

\newpage

\begin{abstract}
We introduce the AIM 2024 UHD-IQA Challenge, a competition to advance the No-Reference Image Quality Assessment (NR-IQA) task for modern, high-resolution photos. The challenge is based on the recently released UHD-IQA Benchmark Database, which comprises 6,073 UHD-1 (4K) images annotated with perceptual quality ratings from expert raters. Unlike previous NR-IQA datasets, UHD-IQA focuses on highly aesthetic photos of superior technical quality, reflecting the ever-increasing standards of digital photography.
This challenge aims to develop efficient and effective NR-IQA models. Participants are tasked with creating novel architectures and training strategies to achieve high predictive performance on UHD-1 images within a computational budget of 50G MACs. This enables model deployment on edge devices and scalable processing of extensive image collections.
Winners are determined based on a combination of performance metrics, including correlation measures (SRCC, PLCC, KRCC), absolute error metrics (MAE, RMSE), and computational efficiency (G MACs). 
To excel in this challenge, participants leverage techniques like knowledge distillation, low-precision inference, and multi-scale training. 
By pushing the boundaries of NR-IQA for
high-resolution photos, the UHD-IQA Challenge aims to stimulate the development of practical models that can keep pace with the rapidly evolving landscape of digital photography. The innovative solutions emerging from this competition will have implications for various applications, from photo curation and enhancement to image compression.
\end{abstract}

\section{Introduction}
\label{sec:intro}


Blind Image Quality Assessment (BIQA) is essential for various applications, including camera benchmarking, professional photo curation, and image enhancement. Despite advances in BIQA models, their effectiveness is constrained by the limitations of existing datasets. Current datasets are primarily annotated at standard definition (SD) resolutions and focus on images with obvious distortions. As a result, BIQA models struggle with high-resolution images that exhibit subtle degradations, which are increasingly common with modern cameras.

These datasets also suffer from a bias toward average or low-quality images, leading to a class imbalance that weakens the generalization of BIQA models. As camera technology advances, producing higher-quality and higher-resolution images, the need for better datasets becomes critical. Moreover, the efficient processing of these high-quality images on edge devices or at scale remains challenging, as most current models are not optimized for such tasks.

We introduce the UHD-IQA challenge as part of AIM 2024 to address these issues. The UHD-IQA benchmark dataset focuses on ultra-high-definition (UHD) images of high aesthetic and technical quality, aiming to fill the gaps in existing benchmarks. The challenge is developing efficient BIQA models that fully leverage this dataset, ensuring high accuracy and computational efficiency for real-world applications.

\subsection{UHD-IQA Benchmark Database}

The dataset comprises 6073 ultra-high-definition (UHD-1, 4K) images, all annotated at a fixed width of 3840 pixels. Unlike existing BIQA datasets, ours focuses on high-quality images with a strong aesthetic appeal, filling a critical literature gap. The images were sourced from Pixabay.com, a repository of CC0-licensed stock photos, and were manually curated to exclude synthetic or heavily edited content. This ensures that the dataset consists of genuine, high-quality photographs. The dataset split is as follows: 4269 for training, 904 for validation, and 900 for testing.

We conducted a crowdsourcing study involving ten expert raters, including photographers and graphic artists, to achieve reliable annotations. Each expert assessed each image at least twice in multiple sessions, yielding 20 ratings per image. The rigorous annotation process and rich metadata, including user and machine-generated tags from over 5,000 categories, provide a comprehensive and reliable resource for training BIQA models.

\begin{figure}[t]
    \centering
    \includegraphics[width=0.8\textwidth]{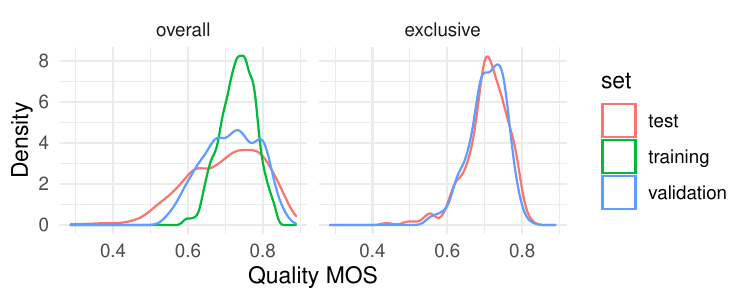}
    \caption{Density of quality MOS per subset. "Overall" includes all image categories, whereas "exclusive" refers to categories that are only part of the validation and test sets.}
    \label{fig:density_plot_dataset}
\end{figure}

Furthermore, the test and validation sets include a special subset of 300 images out of approximately 900 in each set, labeled as \emph{"exclusive"} -- see the MOS density in Fig. \ref{fig:density_plot_dataset}. This subset is selected based on image categories excluded from the training set. The categories for all images were either automatically annotated using AWS Rekognition or manually specified by the image authors when publishing to \url{Pixabay.com}.

The exclusive categories were chosen to be distinct from typical ImageNet ones, focusing on images that do not feature a single dominant object. Instead, they depict multiple scattered objects or wide-spanning scenes. This selection aims to encourage the use of more general-purpose pre-training features. The exclusive categories are Sea, Ocean, Sand, Landscape, Mountain(s), Scenery, City, and Urban.

The performance on the exclusive split also provides valuable insights into each model's generalization capabilities when deviating from the image distribution of the training set.

\subsection{The AIM 2024 Challenge}

The challenge participants were tasked with developing novel BIQA models that efficiently and effectively assess high-resolution images. The proposed models were required to operate below 50 GMACs, ensuring they are lightweight enough for deployment on edge devices or scalable processing. Participants were encouraged to employ strategies such as knowledge distillation and low-precision inference and to select optimal pre-training datasets to meet these requirements.

The challenge was structured around multiple evaluation criteria to determine individual rankings. These criteria included correlation metrics -- Pearson Linear Correlation Coefficient (PLCC), Spearman Rank-order Correlation Coefficient (SRCC), and Kendall Rank Correlation Coefficient (KRCC) -- as well as absolute error metrics such as Mean Absolute Error (MAE) and Root Mean Square Error (RMSE). Additionally, compute efficiency was a critical factor in determining the winning models. By pushing the boundaries of BIQA with this challenge, we aim to drive the development of practical, scalable, and high-performing models that are well-suited for modern, high-quality images.

\paragraph{Associated AIM Challenges.} This challenge is one of the AIM 2024 Workshop\footnote{\url{https://www.cvlai.net/aim/2024/}} associated challenges on: sparse neural rendering~\cite{aim2024snr, aim2024snr_dataset}, UHD blind photo quality assessment~\cite{aim2024uhdbpqa}, compressed depth map super-resolution and restoration~\cite{aim2024cdmsrr}, efficient video super-resolution for AV1 compressed content~\cite{aim2024evsr}, video super-resolution quality assessment~\cite{aim2024vsrqa}, compressed video quality assessment~\cite{aim2024cvqa} and video saliency prediction~\cite{aim2024vsp}.



\section{Proposed Methods}
\label{sec:methods}


Eight methods were submitted for the final round of the challenge. Most solutions consist of ensembles of multiple neural networks, especially Transformer-based~\cite{mehta2021mobilevit, gu2022ntire} models and CLIP-based~\cite{kwon2024clip} models.

As a \emph{baseline}, we propose an efficient solution based on MobileNet V3~\cite{howard2019mobilenet, howard2017mobilenets}. The original high-resolution images are cropped (focusing on the center) at $960 \times 1920$ pixels; these are resized to HD resolution ($1280 \times 720$). Using a fine-tuned MobileNet V3~\cite{howard2019mobilenet} backbone as a feature extractor allows to reduce overfitting and training time and faster inference speed. The baseline model has 3.22 M parameters and a computational complexity of 4.2 GMACs.

\begin{table}[t]
    \centering
    \setlength{\tabcolsep}{10pt} 
    \resizebox{\linewidth}{!}{
    \begin{tabular}{r c c c c c}
        \toprule
        Method	& MAE~$\downarrow$	& RMSE~$\downarrow$ & PLCC~$\uparrow$ & SRCC~$\uparrow$ & KRCC~$\uparrow$ \\
        \midrule
         
        SJTU~(\ref{sec:sjtu})	& \cellcolor{gold} 0.0418	& \cellcolor{silver} 0.0615	& \cellcolor{silver} 0.7985	& \cellcolor{gold} 0.8463	& \cellcolor{gold} 0.6573 \\
        
        GS-PIQA~(\ref{sec:szu}) & \cellcolor{silver} 0.0430	& \cellcolor{gold} 0.0607	& \cellcolor{bronze} 0.7925	& \cellcolor{bronze} 0.8297	& \cellcolor{bronze} 0.6399 \\
        
        CIPLAB~(\ref{sec:ciplab}) & 0.0445	& 0.0638	& \cellcolor{gold} 0.7995	& \cellcolor{silver} 0.8354	& \cellcolor{silver} 0.6419 \\
        
        EQCNet~(\ref{sec:zte})      & \cellcolor{bronze} 0.0438	& \cellcolor{bronze} 0.0621	& 0.7682	& 0.7954	& 0.6055 \\
        
        MobileNet-IQA~(\ref{sec:mobile})	& 0.0463	& 0.0659	& 0.7558	& 0.7883	& 0.5975 \\
        
        NF-RegNets~(\ref{sec:nfregnets})	& 0.0494	& 0.0703	& 0.7222	& 0.7715	& 0.5806 \\

        Challenge Baseline 	& 0.0502	& 0.0733	& 0.6881	& 0.7462	& 0.5537 \\
        
        CLIP-IQA*~(\ref{sec:dominator})	& 0.0519	& 0.0723	& 0.7116	& 0.7305	& 0.5393 \\

        ICL~(\ref{sec:icl})	        & 0.1147	& 0.1364	& 0.5206	& 0.5166	& 0.3615 \\

        \midrule

        HyperIQA \cite{su2020blindly} & 0.070 & 0.118 & 0.103 & 0.553 & 0.389 \\
        
        Effnet-2C-MLSP \cite{wiedemann2023KonxCrossresolutionImageb} & 0.059 & 0.074 & 0.641 & 0.675 & 0.491 \\
        
        CONTRIQUE \cite{madhusudana2022image} & 0.052 & 0.073 & 0.678 & 0.732 & 0.532 \\
        
        ARNIQA \cite{agnolucci2024arniqa} & 0.052 & 0.074 & 0.694 & 0.739 & 0.544 \\
        
        CLIP-IQA+ \cite{wang2023exploring} & 0.089 & 0.111 & 0.709 & 0.747 & 0.551 \\
        
        QualiCLIP \cite{agnolucci2024qualityaware} & 0.066 & 0.083 & 0.725 & 0.770 & 0.570 \\
    
        \bottomrule
    \end{tabular}
    }
    \vspace{2mm}
    \caption{Official \textbf{test} split performance. We highlight the top-3 (gold, silver, bronze) methods for the different metrics. The top section lists methods that participated in the AIM 2024 challenge. The bottom section presents baselines derived from retraining existing methods, which require more than 200 GMACs.}
    \label{tab:test}
\end{table}

\begin{table}[t]
    \centering
    \setlength{\tabcolsep}{10pt} 
    \resizebox{\linewidth}{!}{
    \begin{tabular}{r c c c c c}
        \toprule
        Models	& MAE~$\downarrow$	& RMSE~$\downarrow$ & PLCC~$\uparrow$ & SRCC~$\uparrow$ & KRCC~$\uparrow$ \\
        \midrule
         
        EQCNet~(\ref{sec:zte})	& \cellcolor{gold} 0.0299	& \cellcolor{gold} 0.0383	& \cellcolor{silver} 0.8285	& \cellcolor{silver} 0.8234	& \cellcolor{gold} 0.6342 \\
        
        SJTU~(\ref{sec:sjtu})	& \cellcolor{silver} 0.0318	& \cellcolor{silver} 0.0402	& \cellcolor{gold} 0.8238	& \cellcolor{gold} 0.8169	& \cellcolor{silver} 0.6244 \\
        
        GS-PIQA~(\ref{sec:szu})	& \cellcolor{bronze} 0.0332	& \cellcolor{bronze} 0.0406	& \cellcolor{bronze} 0.8192	& \cellcolor{bronze} 0.8092	& \cellcolor{bronze} 0.6181 \\
        
        CIPLAB~(\ref{sec:ciplab})	& 0.0329	& 0.0423	& 0.8136	& 0.8063	& 0.6143 \\
        
        MobileNet-IQA~(\ref{sec:mobile})& 0.0345	& 0.0439	& 0.7831	& 0.7757	& 0.5824 \\

        NF-RegNets~(\ref{sec:nfregnets})	& 0.0352	& 0.0444	& 0.7968	& 0.7897	& 0.5973 \\

        Challenge Baseline 	& 0.0372	& 0.0482	& 0.7445	& 0.7422	& 0.5504 \\
        
        CLIP-IQA*~(\ref{sec:dominator})	& 0.0398	& 0.0509	& 0.7069	& 0.6918	& 0.5112 \\

        ICL~(\ref{sec:icl})	& 0.0622	& 0.0737	& 0.5217	& 0.5101	& 0.3580 \\

        \midrule
        
        HyperIQA \cite{su2020blindly} & 0.055    & 0.087    & 0.182    & 0.524    & 0.359 \\
        
        Effnet-2C-MLSP \cite{wiedemann2023KonxCrossresolutionImageb} & 0.050 & 0.060 & 0.627 & 0.615 & 0.445 \\
        
        CONTRIQUE \cite{madhusudana2022image} & 0.038 & 0.049 & 0.712 & 0.716 & 0.521 \\
        
        ARNIQA \cite{agnolucci2024arniqa} & 0.039 & 0.050 & 0.717 & 0.718 & 0.523 \\
        
        CLIP-IQA+ \cite{wang2023exploring} & 0.087 & 0.108 & 0.732 & 0.743 & 0.546 \\
        
        QualiCLIP \cite{agnolucci2024qualityaware} & 0.064 & 0.079 & 0.752 & 0.757 & 0.557 \\

        \bottomrule
    \end{tabular}
    }
    \vspace{2mm}
    \caption{Official \textbf{validation} split performance. Comparison of models with top-3 (gold, silver, bronze) highlighted for each metric. The top section lists methods that participated in the AIM 2024 challenge. The bottom section presents baselines derived from retraining existing methods, which require more than 200 GMACs.}
    \vspace{-5mm}
    \label{tab:validation}
\end{table}

\begin{table}[t]
    \centering
    \setlength{\tabcolsep}{10pt} 
    \resizebox{\linewidth}{!}{
    \begin{tabular}{r c c c c c}
        \toprule
        Method	& MAE~$\downarrow$	& RMSE~$\downarrow$ & PLCC~$\uparrow$ & SRCC~$\uparrow$ & KRCC~$\uparrow$ \\
        \midrule

        SJTU~(\ref{sec:sjtu})	& \cellcolor{gold} 0.0292	& \cellcolor{gold} 0.0422	& \cellcolor{gold} 0.6816	& \cellcolor{gold} 0.7407	& \cellcolor{gold} 0.5471 \\
        
        CIPLAB~(\ref{sec:ciplab}) & \cellcolor{silver} 0.0308	& \cellcolor{silver} 0.0439	& \cellcolor{silver} 0.6733	& \cellcolor{silver} 0.7009	& \cellcolor{silver} 0.5078 \\
        
        GS-PIQA~(\ref{sec:szu})	& \cellcolor{bronze} 0.0320	& \cellcolor{bronze} 0.0447	& \cellcolor{bronze} 0.6325	& \cellcolor{bronze} 0.6710	& \cellcolor{bronze} 0.4915 \\

        EQCNet~(\ref{sec:zte})	&  0.0328	&  0.0453	& 0.6227	& 0.6555	& 0.4786 \\
        
        MobileNet-IQA~(\ref{sec:mobile})& 0.0328	& 0.0466	& 0.5916	& 0.5999	& 0.4320 \\

        NF-RegNets~(\ref{sec:nfregnets})	& 0.0338	& 0.0480	& 0.5707 & 0.6099	& 0.4388 \\
        
        CLIP-IQA*~(\ref{sec:dominator})	& 0.0361	& 0.0510	& 0.5113 & 0.5157	& 0.3622 \\

        ICL~(\ref{sec:icl})	& 0.1014	& 0.1138	& 0.4331	& 0.4106	& 0.2802 \\

        \bottomrule
    \end{tabular}
    }
    \vspace{2mm}
    \caption{The performance evaluation of \textbf{exclusive} test split. We highlight the top-3 (gold, silver, bronze) methods for the different metrics.}
    \vspace{-5mm}
    \label{tab:exclusive}
\end{table}

\subsection{Challenge Results}

Table~\ref{tab:test} and Table~\ref{tab:validation} present comparative evaluation results of the eight teams' performance in predicting the quality MOS using various metrics. 

The top three performances for each metric are highlighted, with gold, silver, and bronze representing the first, second, and third-best results, respectively. However, the winner and runner-up teams are ranked considering the final score for each team, which is computed as follows.


\noindent Let \(\mathcal{S}_i\) denote the main score for team \(i\), and $\mathcal{R}(\mathcal{M}_i) = \text{Rank}(\mathcal{M}_i), \mathcal{R}(\mathcal{M}_i) = 1, \dots, N$ is the ranking function that assigns a rank based on the metric value $\mathcal{M}_i$ for each of the $N=8$ participating teams. The best rank is 1. Correlation metrics are ranked highest when they have higher values, whereas absolute errors rank best when they are lowest.

\begin{align}
\mathcal{S}_i = \frac{1}{5} \Big[ & \mathcal{R}\left(\mathcal{M}_i^{\text{MAE}}\right) 
+ \mathcal{R}\left(\mathcal{M}_i^{\text{RMSE}}\right) \notag \\
& + \mathcal{R}\left(\mathcal{M}_i^{\text{KRCC}}\right) 
+ \mathcal{R}\left(\mathcal{M}_i^{\text{PLCC}}\right) \notag 
 + \mathcal{R}\left(\mathcal{M}_i^{\text{SRCC}}\right) \Big]
\end{align}

\noindent \(\mathcal{M}_i^{\text{Metric}}\) represents the value of the previously mentioned metrics.

The team with the lowest main score $\mathcal{S}_i$ is considered to be the winner. Based on the scores obtained and shown in Table 2, team SJTU is the overall competition winner, followed by team SZU SongBai (first runner-up) and team CIPLAB (second runner-up). 

Table~\ref{tab:exclusive} presents a comparative evaluation of the teams' performance in predicting MOS across various evaluation metrics specifically on the exclusive portion of the test set. As expected, the results indicate a noticeable reduction in performance metrics. Interestingly, CIPLAB ranks second in this evaluation (compared to third in the overall ranking in Table \ref{tab:test}), which might be due to better generalization capabilities of the model compared to GS-PIQA. 

\begin{figure}[t]
    \centering
    \includegraphics[width=1\linewidth]{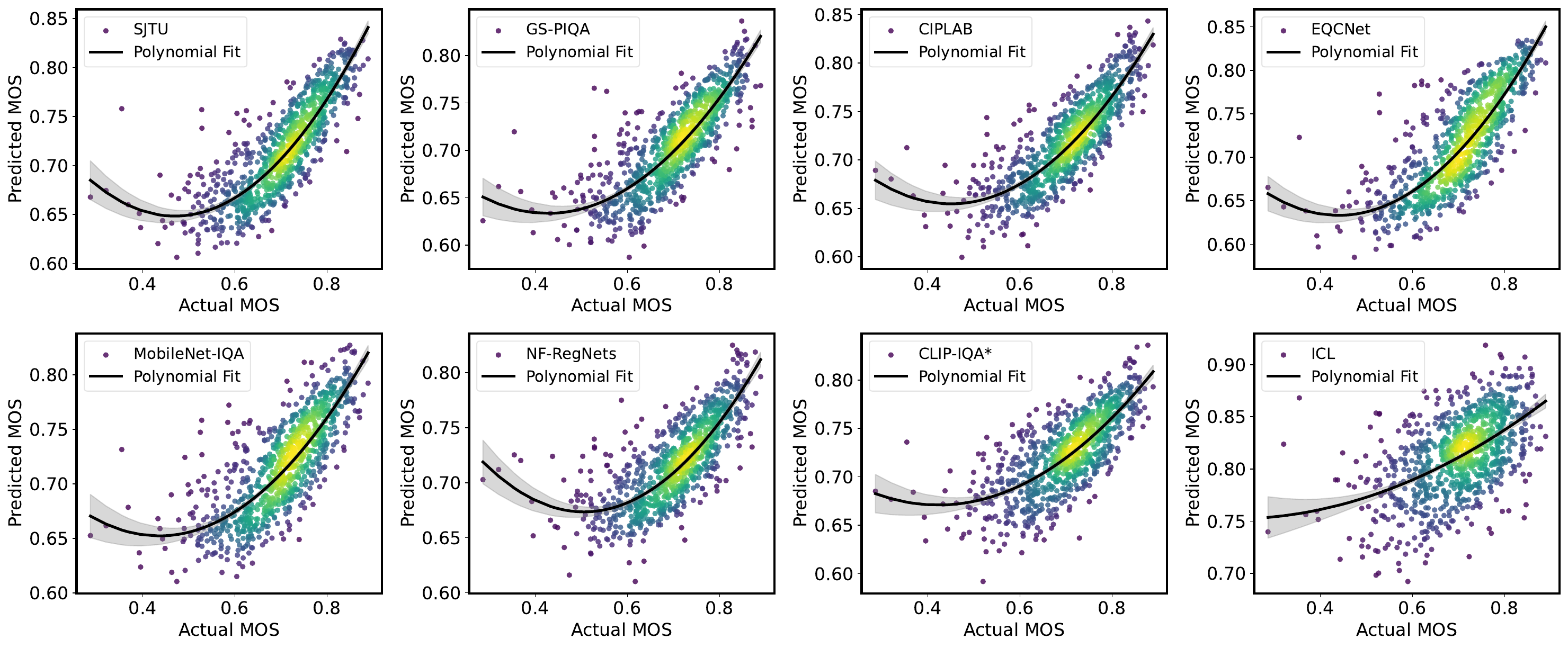}
    \caption{Scatter plots of the predicted quality scores vs ground-truth (actual) MOS. The curves were obtained by a second-order polynomial fitting.}
    \label{fig:perfplots}
\end{figure}

Figure~\ref{fig:perfplots} presents a comparative analysis of predicted quality scores against ground-truth MOS for the eight competing teams. 
Each subplot represents the performance of a particular team, with the team name shown in the legend. The X-axis values represent ground-truth (actual) MOS; the predicted scores are shown on the y-axis. The purple scatter points represent a particular image prediction score, with higher-density areas shown in yellow. The polynomial fit, shown as a black curve, highlights the general trend in the predictions relative to the ground truth.

It can be observed that all teams display a positive correlation between the predicted and ground-truth MOS, as indicated by the upward trend in all subplots. However, the digress of scatter around the fitted curve varies across the subplots indicating the difference in the strength and alignment of this correlation between various teams. For example, the polynomial fit for teams like `\textit{SJTU~(\ref{sec:sjtu})}', `\textit{GS-PIQA~(\ref{sec:szu})}' and `\textit{CIPLAB~(\ref{sec:ciplab})}' show a tighter clustering of data points around the curve, which indicates a better alignment of predicted quality with the ground-truth MOS. On the other hand, teams such as `\textit{ICL~(
ef{sec:icl})}' and `\textit{NF-RegNets~(\ref{sec:nfregnets})}' show more scattered data points. 
Overall, while all teams demonstrate the ability to predict MOS scores with some degree of accuracy, there are clear differences in prediction quality.

\begin{table*}[!ht]
    \centering
    \resizebox{\textwidth}{!}{
    \begin{tabular}{r c c c c c c c}
        \toprule
        Method & Input & \makecell{Training \\  Time (hrs)} & \makecell{Extra\\ Data} & Params. (M) & MACs (G) & GPU  \\
         \midrule
         
         SJTU~(\ref{sec:sjtu}) & $480\times480$  & 12  & Yes & 82.85 & 43.53 & RTX 3090 \\

         GS-PIQA~(\ref{sec:szu}) & $384\times384$  & 4  & No & 144.814 & 50.260 & GTX3090 \\
         
         CIPLAB~(\ref{sec:ciplab}) & $2160\times3840$ & 12  & No & 113 & 44 & RTX 2080 Ti \\

         EQCNet~(\ref{sec:zte}) & $384\!\times\!384$--$1366\!\times\!768$ & 22 & Yes & 30.15 & 12.97 & A800 \\

         MobileViT-IQA~(\ref{sec:mobile}) & $1907\times1231$ & 18 & No & 96.72 & 359.74 & A800 \\ 
         
         MobileNet-IQA~(\ref{sec:mobile}) & $1907\times1231$ & 48 & No  & 81.48 & 46.73 & A800 \\ 

         
         NF-RegNets~(\ref{sec:nfregnets}) & $720\times720$ & $\approx$10 & No & 28.5 & 44.52 & 2$\times$2070 Ti \\
         
         CLIP-IQA*~(\ref{sec:dominator}) & $224\times224$ & 0.25 & No & 151 & 48.5 & A6000 \\
         
         ICL~(\ref{sec:icl}) & $2160\times3840$ & 0.1 & No & 139.1 & 42.09 & A100 \\

         Challenge Baseline & $1280\times720$ & 6 & No & 3.2 & 4.2 & 3090Ti \\
         
         \bottomrule
         
    \end{tabular}
    }
    \vspace{2mm}
    \caption{Training specification for each method. All inputs are 3-channel RGB images; only the spatial dimensions are listed. 
    }
    \vspace{-3mm}
    \label{tab:Training_specification}
\end{table*}

\paragraph{\textbf{Summary of Implementation Details}}

A summary of the methods is provided in Table \ref{tab:Training_specification}, which includes details on the input resolution, computational complexity measured in MACs, and the number of parameters for each model.

In the following sections, we describe the top solutions to the challenge. Please note that the method descriptions were provided by the respective teams or individual participants as their contributions to this report.

\input{teams/sjtu}          

\input{teams/szu}           

\input{teams/CIPLAB}        

\input{teams/ZX_AIE_Vector} 

\input{teams/mobile}        


\input{teams/baseline}      

\input{teams/ma_li}         

\input{teams/icl}           


\section*{Acknowledgements}
This work was partially supported by the Humboldt Foundation. We thank the AIM 2024 sponsors: Meta Reality Labs, KuaiShou, Huawei, Sony Interactive Entertainment, and the University of W\"urzburg (Computer Vision Lab).



\bibliographystyle{splncs04}
\bibliography{refs}

\end{document}

%% file: preamble.tex
\usepackage{graphicx}
\usepackage{comment}
\usepackage{booktabs}
\usepackage{tabularx}
\usepackage{multirow}
\usepackage{xcolor,soul}
\usepackage{subcaption}
\usepackage{caption}
\usepackage{makecell}
\usepackage{adjustbox}

\usepackage{pifont}
\usepackage{xspace}

\definecolor{Gray}{gray}{0.9}
\definecolor{lgray}{gray}{0.95}
\definecolor{LightCyan}{rgb}{0.92,0.92,1}
\definecolor{lyellow}{rgb}{1,1,0.92}
\definecolor{lgreen}{rgb}{0.92,1,0.95}
\definecolor{llblue}{rgb}{0.92,0.93,0.95}
\definecolor{lred}{rgb}{1,0.85, 0.85}
\definecolor{tabhighlight}{HTML}{e5e5e5}


%% file: teams/sjtu.tex
\subsection{Assessing UHD Image Quality from Aesthetics, Distortion, and Saliency}
\label{sec:sjtu}

\emph{Wei Sun,
Weixia Zhang,
Yuqin Cao,
Linhan Cao,
Jun Jia,
Zijian Chen,
Zicheng Zhang,
Xiongkuo Min,
Guangtao Zhai} \\
\textit{Shanghai Jiao Tong University (SJTU), China}

\vspace{5mm}

We design a multi-branch deep neural network (DNN) to evaluate the UHD image quality from three perspectives: \textbf{global aesthetic characteristics, local technique distortions, and salient region perception}, while avoiding direct processing of high-resolution images~\cite{sun2024assessing}. Specifically, a low-resolution image resized from the UHD image, a fragment image composed of local fragments cropped from the equal-size patches of the UHD image, and the center patch cropped from the UHD image are used as inputs to extract the respective features through three branches. The Swin Transformer Tiny~\cite{liu2021swin} pre-trained on the AVA dataset~\cite{murray2012ava} are utilized as the backbone networks of the three branches. The extracted features are concatenated and regressed into quality scores by a two-layer multi-layer perceptron (MLP). We employ the mean square error (MSE) loss and the fidelity loss~\cite{tsai2007frank} to optimize the proposed model. By dividing the overall quality measurement of the high-resolution image into three quality dimension measurements of low-resolution images, our method effectively assesses the quality of UHD images with an acceptable computational complexity. Moreover, we avoid complex model designs and use only the standard DNN structures, making it easy to implement in practical applications and optimize for hardware.

\begin{figure*}[t]
    \centering
    \includegraphics[width=1\linewidth]{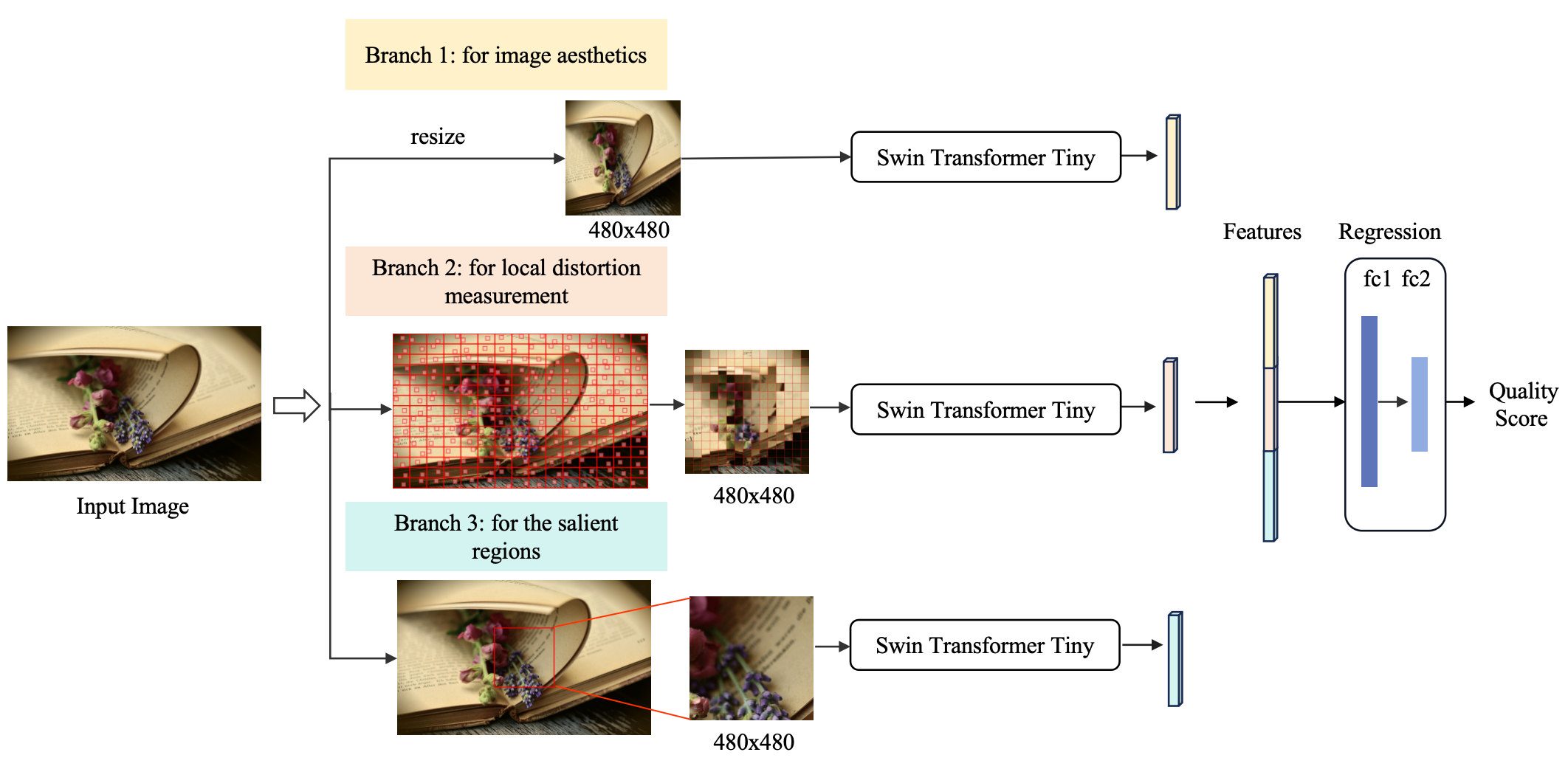}
    \caption{The method proposed by the SJTU Team, using three branches~\cite{sun2024assessing}.}
    \label{fig:sjtu}
\end{figure*}

The proposed model is illustrated in Fig.~\ref{fig:sjtu}. It consists of three branches to extract the quality-aware features from aspects of global aesthetic characteristics, local technique distortions, and salient object perception. 

First, we consider \textbf{image aesthetics}, which encompasses the overall perception of image characteristics such as content, layout, color, contrast, etc. These usually are global features that do not require high resolution. Thus, we resize the UHD image to a low resolution of $480\times480$ and use the low-resolution image as the input of the branch responsible for the aesthetic characteristics. 

Second, we address \textbf{low-level image distortions}, which are typically evident on local image patches and are sensitive to the resolution. We employ a fragment sampling strategy~\cite{wu2022fast}, where the entire image is divided into $15\times15$ equal-sized patches, and a smaller fragment with a resolution of $32\times32$ is randomly cropped from each patch. These fragments are then spliced into a fragment image of $480\times480$, which serves as input to the branch responsible for local distortion measurement. 

Third, since UHD images are often viewed on large screens where the human visual system tends to focus on salient regions, \textbf{the quality of the salient region is crucial for the overall quality}. Considering the center bias of saliency detection~\cite{borji2012state}, we crop the center patch with a resolution of $480\times480$ from the UHD image to extract the quality-aware features for the salient regions.

Finally, we use Swin Transformer Tiny~\cite{liu2021swin} pre-trained on the AVA dataset as the backbone of three branches to extract the corresponding features for each aspect. Note that these three branches do not share the model weights. The extracted features are concatenated as the quality-aware feature representation and then regressed into quality scores via a two-layer MLP network. The two-layer MLP network consists of $128$ and $1$ neurons, respectively. We employ the mean square error (MSE) loss to optimize quality prediction accuracy and the fidelity loss~\cite{tsai2007frank} to optimize quality monotonicity.

%% file: teams/szu.tex
\subsection{Blind Photo Quality Assessment based on Grid Mini-patch Sampling and Pyramid Perception}
\label{sec:szu}

\emph{Songbai Tan~$^1$,
Lixin Zhang~$^2$,
Guanghui Yue~$^2$} \\
\textit{$^1$  School of Management, Shenzhen University, China\\
$^2$ School of Biomedical Engineering, Shenzhen University, China\\
Team SZU
}

\vspace{5mm}

\begin{figure*}[t]
    \centering
    \includegraphics[width=\columnwidth]{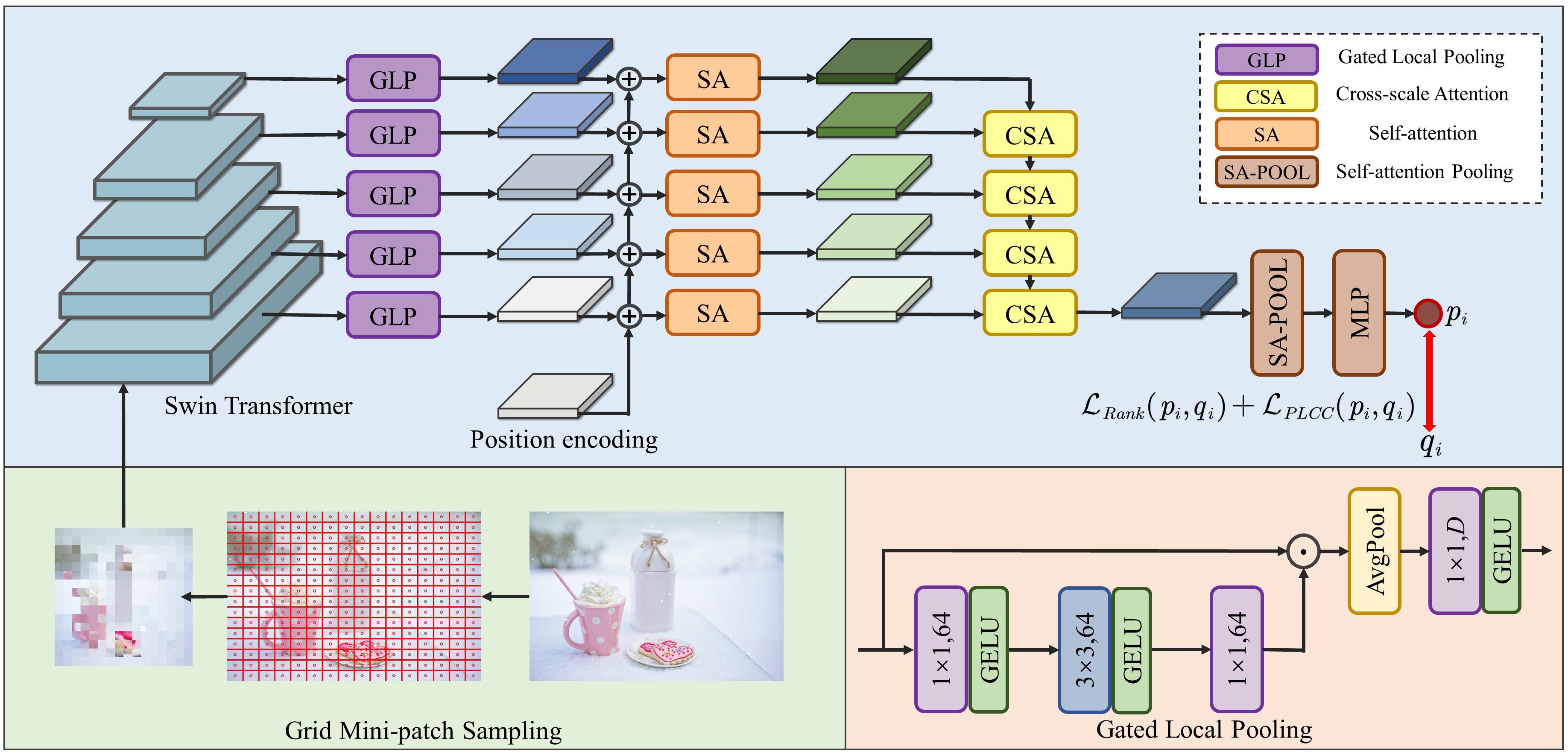}
    \caption{Overview of the proposed GS-PIQA by Team SZU.}
    \label{fig:szu}
\end{figure*}

We propose an effective photo quality assessment method named GS-PIQA, which is an improvement based on CFA-Net\cite{chen2024topiq}. The detailed framework of the model is shown in Fig. \ref{fig:szu}. To enhance the ability to extract global information, we employ the Swin Transformer base network pre-trained on ImageNet as the backbone for GS-PIQA. In addition,  GS-PIQA inherits the gated local pooling (GLP), the self-attention (SA) blocks, and the cross-scale attention (CSA) blocks in CFA-Net to enhance the multi-scale features across different layers. Through this top-down feature extraction and enhancement method, the model can form a pyramid perception capability. Given the high resolution of photos, directly resizing them would result in a significant loss of quality-related information. The common approach is to perform multiple crops on the image and predict the quality for different cropped regions, averaging these regional quality scores to obtain the overall quality. While this method avoids the distortion and information loss caused by resizing, the small cropped areas can only represent local information, leading to substantial bias in overall quality prediction. To address these issues, we adopt the grid mini-patch sampling method for high-resolution images, which reduces the input resolution while preserving the semantic and quality features of the original image.  Specifically, we cut the input high-resolution image $\mathcal{P}$ into a uniform grid of $N\times N$, representing them as $G=\{g_{(0,0)},...,g_{(i,j)},...,g_{(N-1,N-1)}\}$, where $i$ and $j$ indicate that the grid is in the $i$-th row and $j$-th column, respectively. For each grid $g_{(i,j)}$, we randomly take a small region of size $n\times n$ and splice all the obtained small regions to obtain the final sample image of size $K\times K$. In this experiment, the values of $N$, $n$, and $K$ are set to 16, 24, and 384, respectively. The uniform grid mini-patch sampling process is formalized as follows:

\begin{equation}
    g_{(i,j)}=\mathcal{P}[\frac{i\times H}{N}:\frac{(i+1)\times H}{N},\frac{j\times W}{N}:\frac{(j+1)\times W}{N}],
\end{equation}

where $H$ and $W$ are the height and width of the input image respectively. The detail information of GS-PIQA is illustrated in Table. \ref{tab:Training_specification}.

During the training process, we randomly sampled an image 10 times. The average of the quality prediction results of 10 samples is taken as the final quality prediction result of the image. To train the network, we employed the Rank and PLCC loss functions, which can be expressed as follows:

\begin{equation}
    \mathcal{L}=\mathcal{L}_{\text{Rank}}(p_i, q_i)+\mathcal{L}_{\text{PLCC}}(p_i, q_i)
\end{equation}

where $p_i$ and $q_i$ represent the predicted and true scores, respectively. Since the predicted results are not in the same range as the true quality scores, we map the predicted results as follows:

\begin{equation}
    p_i=\frac{p_i-\text{min}(p_i)}{\text{max}(p_i)-\text{min}(p_i)}\times (\text{max}(q_i)-\text{min}(q_i)) + \text{min}(q_i)
\end{equation}

\subsubsection{Implementation details}
We trained and tested only on the UHD-IQA database and divided the database into training and test sets according to 8:2. The input images were processed using the grid mini-patch sampling method to obtain samples of size $384\times384$. To train GS-PIQA, we used the AdamW optimizer, initializing the learning rate at $10^{-4}$ and the weight decay coeffective at $10^{-5}$. The network was trained for 10 epochs with the cosine learning rate decay strategy, setting the temperature coefficient $T$ to 5. 

The training process was divided into two phases. The first phase was trained using the above configuration, saving the results that performed best in the test set. In the second phase, we loaded the weights from the first phase and only fine-tuned the last fully connected layer. We increased the number of samples per image in the training set to 30. The fine-tuning learning rate was set to $5\times 10^{-5}$, with a weight decay of $10^{-5}$, for training 10 epochs.

%% file: teams/CIPLAB.tex
\subsection{High Resolution Patch Based Transformer with Quality-aware Feature Extraction}
\label{sec:ciplab}

\emph{Daekyu Kwon, Dongyoung Kim, Seon Joo Kim} \\
\textit{CIPLAB, Yonsei University, Korea}

\vspace{5mm}


We propose a Vision Transformer~\cite{dosovitskiy2021an} based IQA method which can efficiently handle arbitrary high-resolution images, using high-resolution patch strategy and quality-aware CNN extractor~\cite{kwon2024clip}.
When applying the conventional ViT architecture to UHD images, excessive computation is required due to the large number of patches needed for training.
To address this issue, we propose an architecture that can efficiently compute with fewer patches for high-resolution images by increasing the patch size, typically around 12 or 14, to 224.
Doing so enables us to effectively handle UHD images with Vision Transformer architecture less than 50G MACs.

Furthermore, by employing high-resolution patches, we integrate an advanced CNN that can extract more meaningful features for IQA in the patch projection stage with ViT rather than the simple CNN utilized by the conventional ViT architecture.
We first train a CNN-based feature extractor through a quality-aware pre-training method and utilize it as a feature extractor at the fine-tuning stage.    

\begin{table}
    \centering
    \setlength{\tabcolsep}{8pt}
    \begin{tabular}{l c c}
         \toprule
         Method & SRCC & PLCC  \\
         \midrule
         MobileNet(ImageNet-21k) + ViT & 0.7828 & 0.7860 \\
         MobileNet(ATTIQA) + ViT & 0.8063 & 0.8136 \\
         \bottomrule
    \end{tabular}\vspace{5pt}
    \caption{Comparison of CIPLAB ensemble results using Quality-Aware CNN Extractor. We measure SRCC and PLCC using the official validation set.}
    \label{tab:results}
\end{table}


\begin{figure*}[t]
    \centering
    \includegraphics[trim={0 0 15mm 0},width=0.9\textwidth]
    {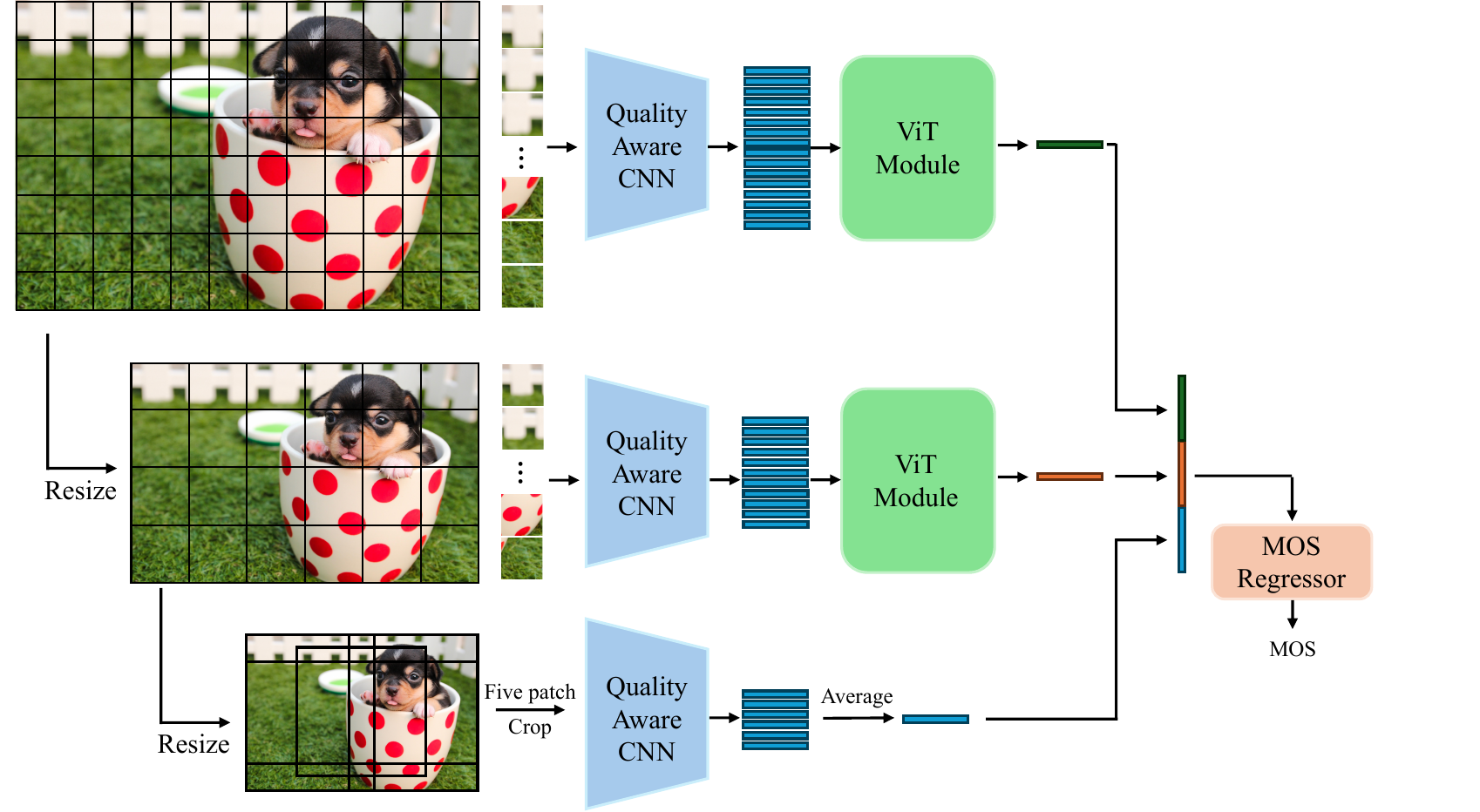}
    \caption{The overall process of the CIPLAB ensemble method. We utilize three types of various sized images. First, we patchify each image and encode them into features using a pre-trained quality-aware CNN. The features extracted from high-resolution images are encoded by a ViT module, while the features extracted from low-resolution images are averaged. We then concatenate these features and predict the MOS using a 2-layer MLP regressor.}
    \label{fig:ciplab}
\end{figure*}

\subsubsection{Global Method Description}

Our method consists of two primary stages: a pre-training stage (where we only train a CNN-based feature extractor) and a fine-tuning stage.

Our model consists of two primary components: a CNN-based feature extractor and a transformer-based feature aggregation module.
For the CNN-based feature extractor, we employ MobileNet-v3-large~\cite{howard2019mobilenet} from the timm library as the backbone and attach 2-layer MLPs to each attribution head, following the ATTIQA approach.
For the transformer-based feature aggregation module, we utilize the default Vision Transformer architecture as a backbone, adopting Global Average Pooling for the final feature extraction instead of the CLS token.
As we mentioned, we use a patch size of 224 and encode each patch into features using the CNN-based feature extractor. 

\textbf{Pre-training Stage.}  Our pre-training method is derived from ATTIQA~\cite{kwon2024clip}.  Due to computational restrictions, we train MobileNet-V3 as a lightweight backbone using ATTIQA's pre-training strategy with ImageNet-21k.  We note that all pre-training setups are identical to ATTIQA's setup, and additional details are provided in Section 3.

\textbf{Fine-tuning Stage.} Inspired by MUSIQ~\cite{ke2021musiq}, we also utilize a multi-scale input strategy. To implement this strategy, we use three types of inputs: (a) the original resolution image (\textit{W}=3840 ), (b) a 1/4 resolution image (\textit{W}=960 ), and (c) a tiny resolution image (\textit{W}=256 ). 
Each image is encoded into features independently using different CNN-based feature extractors.

Given that high-resolution images ((a) and (b)) are sufficient to integrate with the transformer, we extract features of high-resolution images using the transformer with images (a) and (b).
For image (c), we compute the final feature by extracting five features for each side and center crop and averaging them.
After extracting three features for high-resolution images and low-resolution image, we concatenate them into one feature and predict ground truth MOS using 2-layer MLP.


\begin{table}[]
    \centering
    \setlength{\tabcolsep}{8pt}
    \begin{tabular}{lcc}
        \toprule
        \textbf{Details}            & \textbf{Pre-train Stage}       & \textbf{Finetune Stage} \\ 
        \midrule
        \textbf{Backbone}           & MobileNetV3                   & MobileNetV3 + Vision Transformer \\ 
        \textbf{Loss}               & MarginRankingLoss             & L1 Loss \\ 
        \textbf{Optimizer}          & AdamW                         & AdamW \\ 
        \textbf{Learning Rate}      & 1e-4                          & 1e-5 \\ 
        \textbf{GPU}                & 8 $\times$ V100                      & 4 $\times$ RTX2080 Ti \\ 
        \textbf{Dataset}            & Imagenet 21k                  & UHD-IQA Dataset \\ 
        \textbf{Times}              & 4d                            & 10h \\ 
        \textbf{Augmentation}       & RandomResizedCrop             & RandomHorizontalFlip(p=0.5) \\ 
        \bottomrule
    \end{tabular}
    \vspace{5pt}
    \caption{Implementation details for Pre-train and Finetune Stages of CIPLAB.}
\end{table}

%% file: teams/ZX_AIE_Vector.tex
\subsection{Learning from Strong to Weak, Enhanced Quality Comparison Network via Efficient Transfer Learning}
\label{sec:zte}

\emph{Yunchen Zhang,
Xiangkai Xu,
Hong Gao,
Yiming Bao,
Ji Shi,
Xiugang Dong,
Xiangsheng Zhou,
Yaofeng Tu} \\
\textit{ZTE Corporation}

\vspace{5mm}

We propose two IQA models with different parameter scales. The teacher model, called Ensemble IQANet (EIQANet), is a large-parameter model designed to explore the upper bound of performance on UHD datasets\cite{hosu2024uhd}. The student model, Enhanced QCNet (EQCNet), is based on geometric order learning \cite{lee2022geometric} for accurate rank estimation, serving as a lightweight model to meet the requirements of real-time applications. It is worth noticing that a significant performance gap lies in EIQANet and EQCNet. 
Furthermore, we designed a multi-stage knowledge transfer strategy involving three training steps: pre-training, fine-tuning, and calibration. 
This approach facilitates effective knowledge transfer between heterogeneous models and drives the construction of a well-arranged, well-clustered embedding space.

\begin{table}[h]
    \centering
    \begin{tabular}{l c c c c}
         \toprule
         Method & KRCC & SROCC & RMSE & MAE \\
         \midrule
         Q-Align \cite{wu2023q} & 0.3069 & 0.4412 & 0.1685 & 0.0289 \\
         Q-Align-LoRA (finetuned) \cite{wu2023q} & 0.2052 & 0.2624 & 0.0748 & 0.0597 \\
         Compare2Score \cite{zhu2024adaptive} & 0.2553 & 0.3735 & 0.1651 & 0.1524 \\
         QCN \cite{shin2024blind} & 0.2977 & 0.42756 & 0.0615 & 0.0496 \\
         QCN-UHD (finetuned) & 0.4707 & 0.6485 & 0.0581 & 0.0484 \\
         EQCN (Ours) & \textbf{0.6520} & \textbf{0.8403} &\textbf{ 0.0371} & \textbf{0.0289} \\
         \bottomrule
    \end{tabular}
    \vspace{2mm}
    \caption{Performance Comparisons of EQCN and latest BIQA methods.}
    \vspace{-5mm}
    \label{tab:results_zte}
\end{table}


\begin{figure*}[t]
    \centering
    \includegraphics[width=\linewidth]{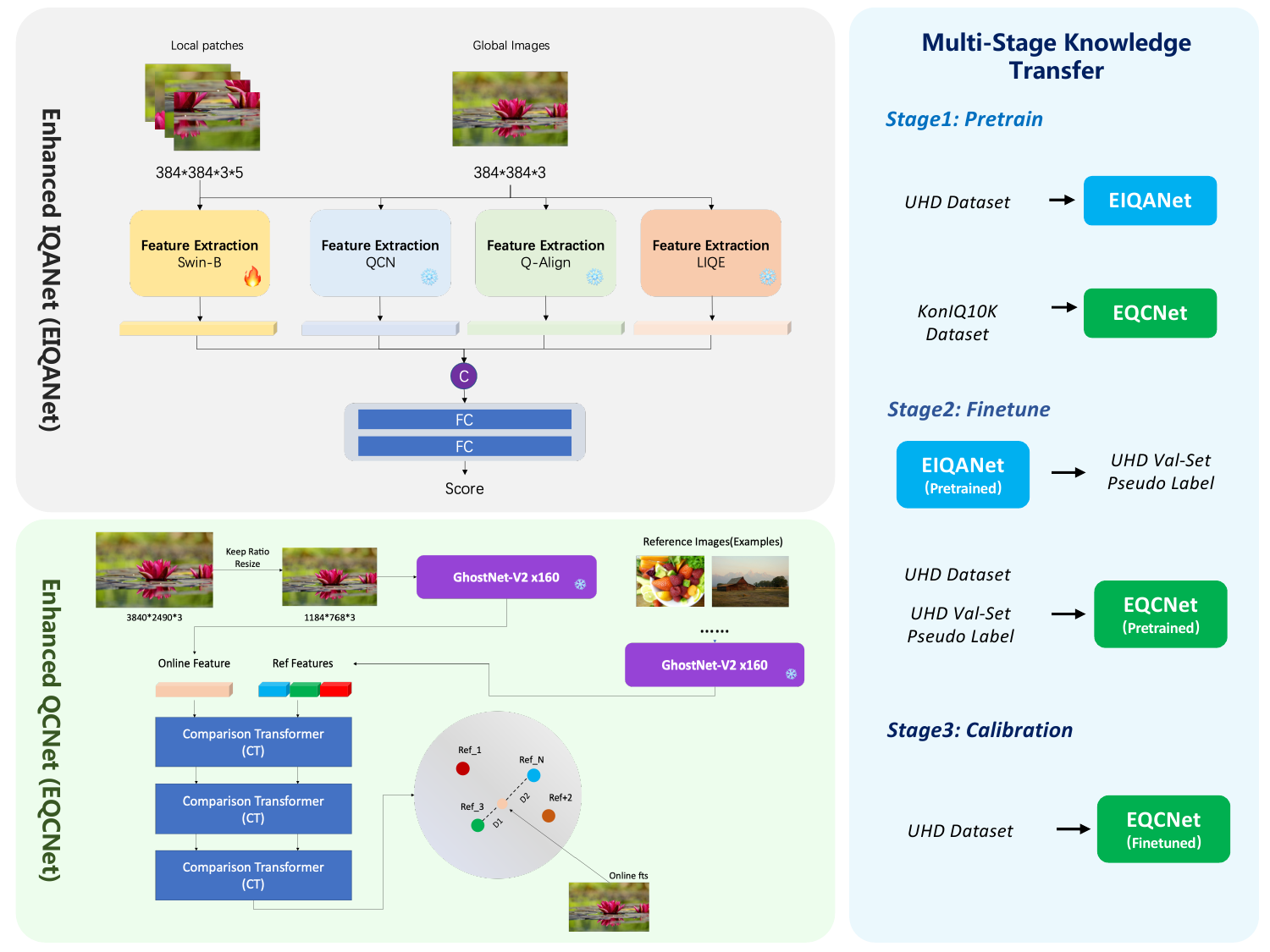}
    \caption{Illustration of Enhanced IQANet (EIQANet), Enhanced Quality Comparison Network (EQCNet) and Multi-Stage Knowledge Transfer Strategy.}
    \label{fig:zte}
\end{figure*}

\subsubsection{Enhanced IQANet}
\label{sec:Enhanced IQANet}
Inspired by RD-VQA \cite{sun2024enhancing}, we propose the Enhanced IQANet (EIQANet). Given that the image resolutions in the UHD dataset all exceed 2K, we have made several advancements in image processing and feature extraction to fully utilize the information in high-resolution images. To better focus on the objective evaluation metrics of IQA tasks, we have also refined the loss functions. Our approach includes the following improvements:

\textbf{High-Resolution Image Processing.} 
The latest VLM model \cite{chen2024far} processes images up to 4K resolutions without altering the image feature encoder architecture. 
We introduce the dynamic patch-slicing mechanism that allows a high-resolution image to be divided into up to 4 patches, capturing high-resolution features.

\textbf{Multi-Model Feature Fusion.} To boost the performance of the BIQA network, we introduce several advanced IQA models to provide auxiliary features:

QCN \cite{shin2024blind}: As the first image quality prediction model based on geometric order learning \cite{lee2022geometric}, QCN extracts features with strong generalization performance.

Q-Align \cite{wu2023q}: As a VLM model, Q-Align leverages powerful LLMs to offer highly interpretable image quality assessments. We utilize the penultimate layer embeddings as features.

LIQE \cite{zhang2023blind}: Based on image-text contrastive learning, the LIQE image encoder provides rich image features aligned with natural language.

Similar to RD-VQA, we employ an offline feature extraction method to obtain the above auxiliary features.

\textbf{Refined Loss Function.} \cite{shin2024blind} considered only the $l1$ loss function during training, which overlooked the ordered sequence relationship of image quality within a batch, resulting in sub-optimal performance on objective evaluation metrics like PLCC. To address this, we additionally incorporate PLCC and SRCC loss functions, enabling the network to consider the absolute scores of the current samples and the relative order of image quality assessments within a batch.

\subsubsection{Enhanced QCNet (EQCNet)}
The proposed EIQANet significantly improves performance metrics on the UHD dataset \cite{hosu2024uhd}. However, EIQANet's reliance on offline feature extraction and its large number of parameters severely limit its practicality in real-world scenarios.

To address these limitations, following the design of \cite{shin2024blind}, we introduce the comparison transformer (CT) to map each instance into a feature vector in an embedding space. Furthermore, the geometric order learning (GOL) \cite{lee2022geometric} uses the reference points to satisfy both order and metric constraints and construct a well-arranged embedding space.


\textbf{Efficient Backbone Design.} To ensure computational efficiency when processing high-resolution images, we use the GhostNetV2 \cite{tang2022ghostnetv2} as the backbone for image feature extraction. GhostNetV2, benefiting from the DFC attention mechanism \cite{tang2022ghostnetv2} and depth-wise separable convolutions, ensures both feature diversity and model efficiency. We believe that GhostNetV2's efficiency in modeling image features ensures that even after the features are projected through GOL, they retain their discriminative power. 


\subsubsection{Multi-Stage Knowledge Transfer} 
Based on \cite{lee2022geometric}, the GOL method is significantly influenced by the initialization of reference points, which depend on the distribution characteristics of the provided training dataset. However, a substantial distribution difference exists between the original UHD training set and the test set data \cite{hosu2024uhd}.

To address this, we designed a multi-stage knowledge transfer method. 
First, we pre-trained EQCNet using the KonIQ-10k \cite{hosu2020koniq} dataset to impart an initial image quality perception capability. Second, we utilized EIQANet, as mentioned in Sec. \ref{sec:Enhanced IQANet}, to generate pseudo-labeled data on the validation set of the UHD dataset \cite{hosu2024uhd}. This pseudo-labeled data was then combined with the UHD training set for joint fine-tuning.
Third, we fine-tuned EQCNet to align its embedding space with the joint UHD dataset distribution. Notably, EQCNet was initialized with weights from the model pre-trained on the KonIQ-10k dataset. Finally, recognizing potential noise and errors in the pseudo-labels, we further calibrated the EQCNet model using the UHD training set to obtain the final model.

This training method mitigates the slow convergence issue of small-parameter models and transfers knowledge from large-parameter models through a progressive learning strategy. This approach guides the EQCNet in learning a comprehensive feature mapping space, enhancing the performance and robustness of the BIQA model.

\subsubsection{Additional Implementation details}
We implemented EIQANet and EQCNet using PyTorch. For EIQANet, we used the Adam optimizer with a learning rate of $10^{-5}$ during the pre-training stage. Additionally, we trained the model 10 times and averaged the results to achieve robust score predictions.

The training strategy for EQCNet is more complex. During the pre-training stage, we used the AdamW optimizer with a learning rate of $5\times10^{-5}$ and trained the model for 100 epochs on the KonIQ-10k dataset. For the fine-tuning and calibration stages, we switched to the Lion optimizer, setting the learning rates to $3\times10^{-5}$ and $5\times10^{-5}$, respectively. The fine-tuning stage consisted of 100 epochs on the mixed dataset, while the calibration stage was limited to 20 epochs on the UHD train set.



%% file: teams/mobile.tex
\subsection{MobileIQA: No-Reference Image Quality Assessment for Mobile Devices using Teacher-Student Learning}
\label{sec:mobile}

\emph{Zewen Chen~$^{1,2}$,
Shunhan Xu~$^3$,
Haochen Guo$^4$,
Yun Zeng$^{5}$,
Shuai Liu$^3$,
Jian Guo$^6$,
Juan Wang$^1$,
Bing Li$1$, 
Dehua Liu$^7$ and Hesong Liu$^7$} \\
\textit{$^1$ State Key Laboratory of Multimodal Artificial Intelligence Systems, CASIA\\
$^2$ School of Artificial Intelligence, University of Chinese Academy of Sciences\\
$^3$ College of Smart City, Beijing Union University \\
$^4$ College of Information and Electrical Engineering, Hebei University \\
$^5$ School of Economics and Management, China University of Petroleum-Beijing \\
$^6$ College of Robotics, Beijing Union University \\
$^7$ SHANGHAI TRANSSION INFORMATION TECHNOLOGY LIMITED \\}

\vspace{5mm}

To address the challenge of high-resolution image quality assessment, we explore a structure based on MobileViT\cite{mehta2021mobilevit} and MobileNet\cite{howard2017mobilenets} as backbone networks, namely MobileViT-IQA and MobileNet-IQA~\cite{chen2024mobileiqa}. Inspired by the multiple scores given by human annotators, we designed a multi-view opinion (MVO) module. This module can fuse the features extracted by the backbone network, simulating the assessment opinions of different annotators, and ultimately integrate them into an image quality score.

When dealing with high-resolution images, two challenges arise: (1) MobileViT demonstrates excellent performance but has high MACs, making it difficult to deploy on mobile devices; (2) MobileNet offers high computational efficiency, but its performance is not as robust as MobileViT. To address these issues, we employ knowledge distillation\cite{chen2022teacher}. We first train a high-performance MobileViT-IQA model and then use it as a teacher model to guide the learning of the MobileNet-IQA. This model supports outputs with resolutions up to $1907\times1231$ and requires only about 49 GMACs.

This approach effectively balances high performance and computational efficiency, providing a viable solution for high-resolution image quality assessment on mobile devices.

\subsubsection{Model Design}
 We take the features captured from five layers in the MobileViT and MobileNet. Many existing works prove that the multi-layer features are helpful for the IQA task\cite{chen2022teacher, chen2024promptiqa, wang2023hierarchical, Su_2020_CVPR}.

\begin{figure*}[t]
    \centering
    \includegraphics[width=\textwidth]{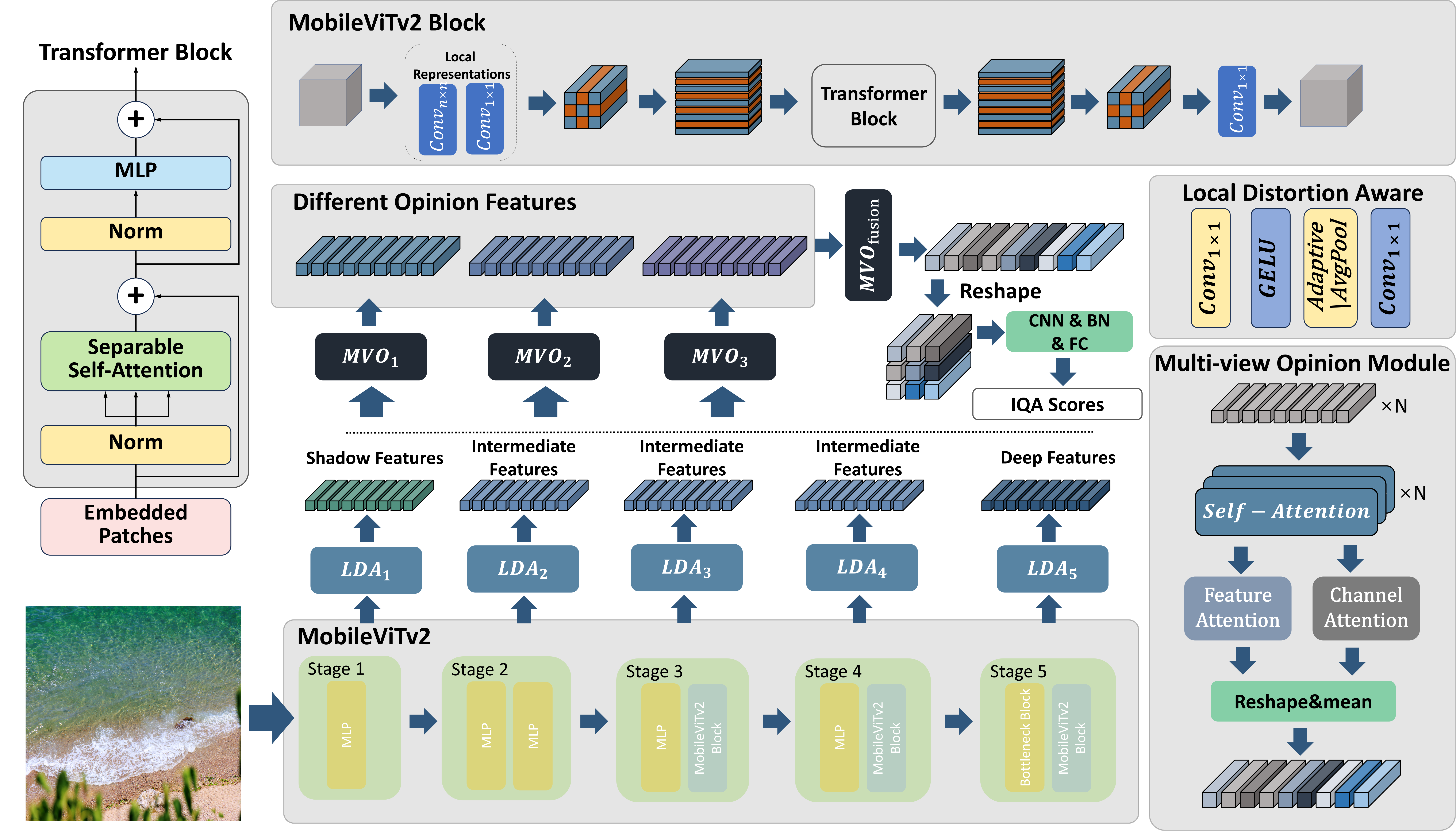}
    \caption{Framework of the teacher model MobileViT-IQA~\cite{chen2024mobileiqa}. The student model MobileNet-IQA shares the same framework, with MobileNet as its backbone.}
    \label{fig:teacher}
\end{figure*}

The teacher model (MobileViT-IQA) is shown in Fig~\ref{fig:teacher}. First, multi-scale features are extracted from five layers of MobileViT, enabling the model to comprehend image quality more comprehensively. Subsequently, these features are fused and dimensionally reduced through a Local Distortion Aware (LDA) module. The processed five features are then input into three Multi-view Opinion (MVO) modules with different weight initializations, generating three distinct opinion features that simulate subjective opinions of the same image by multiple assessors. Finally, these three opinion features are integrated through an additional MVO module, followed by reshaping, convolutional neural network (CNN), and fully connected (FC) layer operations to derive the final image quality score.
The student model (MobileNet-IQA) shares the same framework as MobileViT-IQA but uses MobileNet as the backbone.

\begin{figure*}[t]
    \centering
    \includegraphics[width=\textwidth]{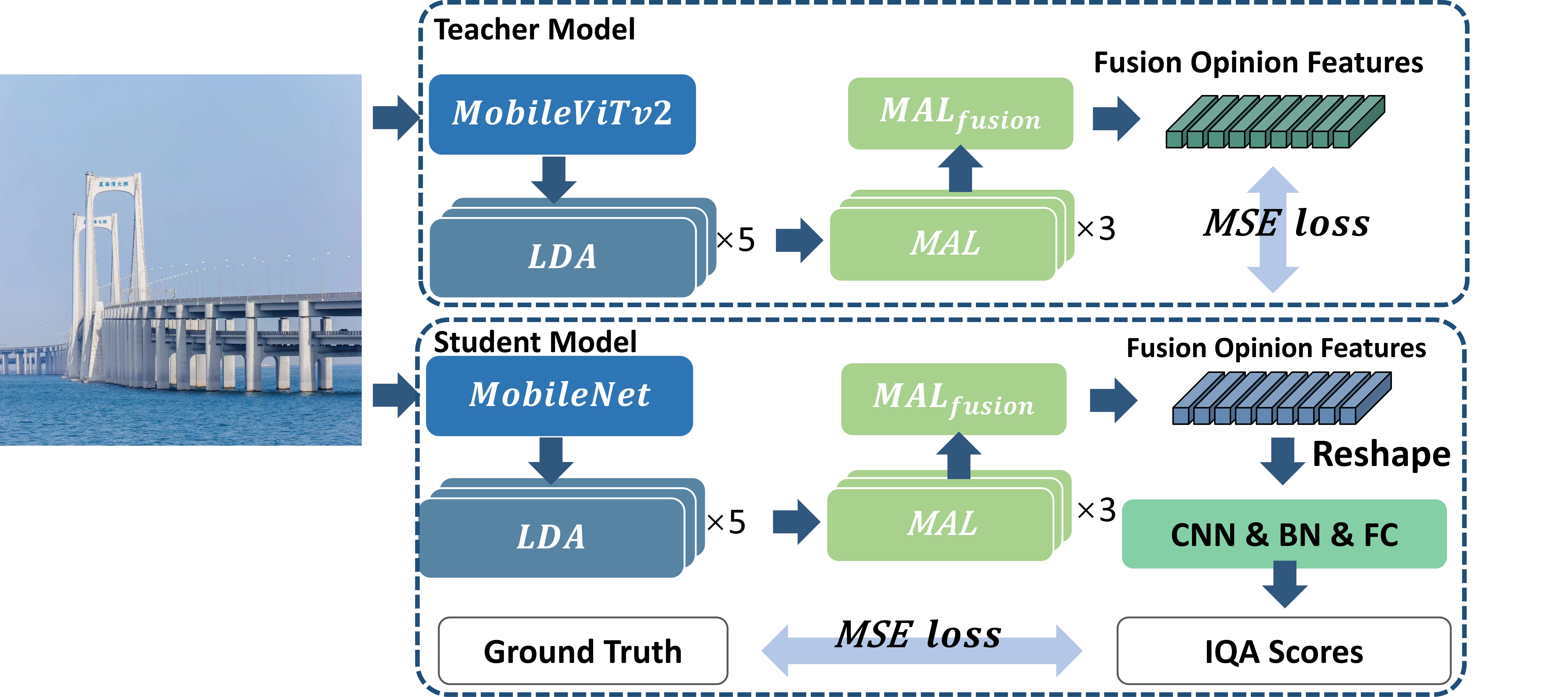}
    \caption{Model distillation process of MobileNet-IQA. The purpose of teacher-student learning is achieved by supervising the Different Opinion Features of the teacher and student networks.}
    \label{fig:distill}
\end{figure*}

The distillation process is shown in Fig~\ref{fig:distill}. Since the MobileViT-IQA and the MobileNet-IQA share the same framework, distilling the teacher's knowledge to the student is more efficient. We take the $MSE$ loss to supervise the discrepancy between the Different Opinion Features (DOF) from the teacher and student models. During training, the discrepancy between the predicted and GT scores is also supervised by the $MSE$ loss.

\subsubsection{Multi-view Opinion}
The motivation is that individuals often have diverse subjective perceptions and regions of interest when viewing the same image. To this end, we employ multiple MVOs to learn attention from different viewpoints.
Each MVO is initialized with different weights and updated independently to encourage diversity and avoid redundant output features. The number of MVOs can be flexibly set as a hyper-parameter. In this work, we set the number to 3. 
As shown in Fig~\ref{fig:teacher}, the MVO starts from $N$ self-attentions (SAs), each of which is responsible for processing a basic feature $\mathbf{f}_j$ ($1\leq j \leq N$). 
The outputs of all the SAs are concatenated, forming a multi-level aggregated feature $\mathbf{F}\in \mathbb{R}^{C\times D \times N}$. Then $\mathbf{F}$ passes through two branches, i.e., a pixel-wise SA branch and a channel-wise SA branch, which apply an SA across spatial and channel dimensions, respectively, to capture complementary non-local contexts and generate multi-view attention maps.
In particular, for the channel-wise SA, the feature $\mathbf{F}$ is first reshaped and permuted to convert the size from $C\times D \times N$ to $D\times (C \times N)$. After the SA, the output feature is permuted and reshaped back to the original size $C\times D \times N$. Subsequently, the outputs of the two branches are added and average pooled, generating an opinion feature.
The design of the two branches has two key advantages. 
First, implementing the SA in different dimensions promotes diverse attention learning, yielding complementary information. Second, contextualized long-range relationships are aggregated, benefiting global quality perception.

\subsubsection{Image Quality Score Regression.}
Assuming that $M$ opinion features are generated from $M$ MVOs. To derive a global quality score from the collected opinion features, we utilize an additional MVO. The MVO integrates diverse contextual perspectives, resulting in a comprehensive opinion feature that captures essential information. This feature is then processed through a transformer block, three convolutional layers with kernel sizes of $5 \times 5$, $3 \times 3$, and $3 \times 3$ to reduce the number of channels, followed by two fully connected layers that transform the feature size from 128 to 64 and from 64 to 1. Finally, we obtain a predicted quality score.

\subsubsection{Additional Implementation Details}
We use the MSE loss to reduce the discrepancy between predicted and GT scores. Then, we use the Adam optimizer with a learning rate of $10^{-5}$ and a weight decay of $10^{-5}$. The learning rate is adjusted using the Cosine Annealing for every 50 epochs. We train the teacher model for 100 epochs (about 18h) with a batch size of 4 and the student model for 300 epochs (about 48h) with a batch size of 8 on one NVIDIA RTXA800. 

%% file: teams/baseline.tex
\subsection{Multi-scale NF-RegNets Ensemble}
\label{sec:nfregnets}

\emph{Grigory Malivenko} \\

\vspace{3mm}

The solution contains three parts, as well as the fusion block. Each sub-model is a NFRegNet~\cite{brock2021highperformancelargescaleimagerecognition} (Norm-Free RegNet) model (\textit{nf-regnet-b1}) trained to predict photo quality on a specific resolution (1:1, 1:2, and 1:3). Features of these models are being fused together and used for the final photo quality estimation.

Without TTAs (test-time augmentations), it takes only 19.08 GMACs to process a photo. Each sub-model takes around 6.36 GMACs to run, and the fusion/classification block takes 0.74 MMACs. This fact makes it possible to perform TTAs very effectively: calculate features for all sub-models separately and then use the fusion/classification block for each possible combination. The runtime is 40 ms for each photo with TTAs, and 15 ms without TTAs.

\begin{figure*}[t]
    \centering
    \includegraphics[width=1\linewidth]{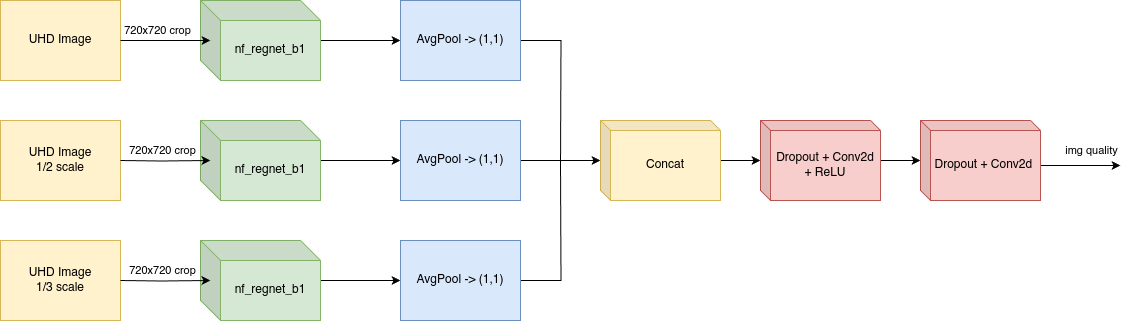}{}
    \caption{Multi-scale NF-RegNets ensemble solution.}
    \label{fig:nfregnet}
\end{figure*}

\paragraph{Implementation details}

PyTorch with Adam optimizer was used. Standard learning rate $10^{-3}$ with step lr-scheduler was used (every 15 steps, factor 0.8). Every sub-model was trained for 150 epochs. Then, after merging sub-models into a single model, only the fusion block was trained for 20 epochs (while sub-model weights were frozen). Finally, the whole model was fine-tuned for another 20 epochs.

Each sub-model training process took around 2 hours, and initial fusion block training took around 30 minutes. The whole model needed to be fine-tuned for another 2 hours. Only random crops and random flips were used for augmentation.

For the final version of the solution, the model was trained on a whole dataset and fine-tuned on a pseudo-labeled validation part.


%% file: teams/ma_li.tex
\subsection{Hybrid Local-Global Image Quality Assessment}
\label{sec:dominator}

\emph{Xingyuan Ma, Cheng Li} \\

\vspace{5mm}

We divide the original image into several patches and score them separately. To avoid the impact of image content on model performance, we randomly disrupt the order of the above patches and reassemble new images for scoring. Finally, the scores of the original image, several patches, and the reorganized new image are averaged to create the final score.

\subsubsection{Global Method Description}
Our method, denoted as CLIP-IQA*, is based on CLIP-IQA. Unlike CLIP-IQA, we use positional encoding, and the model's input is fixed to $224 \times 224$ pixels.

The prompts we used are 'The quality of this photo is bad', 'The quality of this photo is poor', 'The quality of this photo is fair', 'The quality of this photo is good', 'The quality of this photo is perfect'. 

In this challenge, an image resolution that is too large will require considerable calculation. A common approach is to downsample the original image to a very small resolution, such as $224 \times 224$, resulting in a severe loss of input information. In addition, this method violates the logic of subjective image quality evaluation. Inspired by the process of image quality evaluation from the whole to the part or from the part to the whole, we divide the original image into several patches and score them separately. The dimensions of image quality evaluation are related to noise, clarity, color, details, etc. To avoid the impact of image content on model performance, we randomly disrupt the order of the above patches and reassemble new images for scoring. Finally, the scores of the original image, several patches, and the reorganized new image are averaged as the final score.

\begin{figure*}[t]
    \centering
    \includegraphics[width=1.0\linewidth]{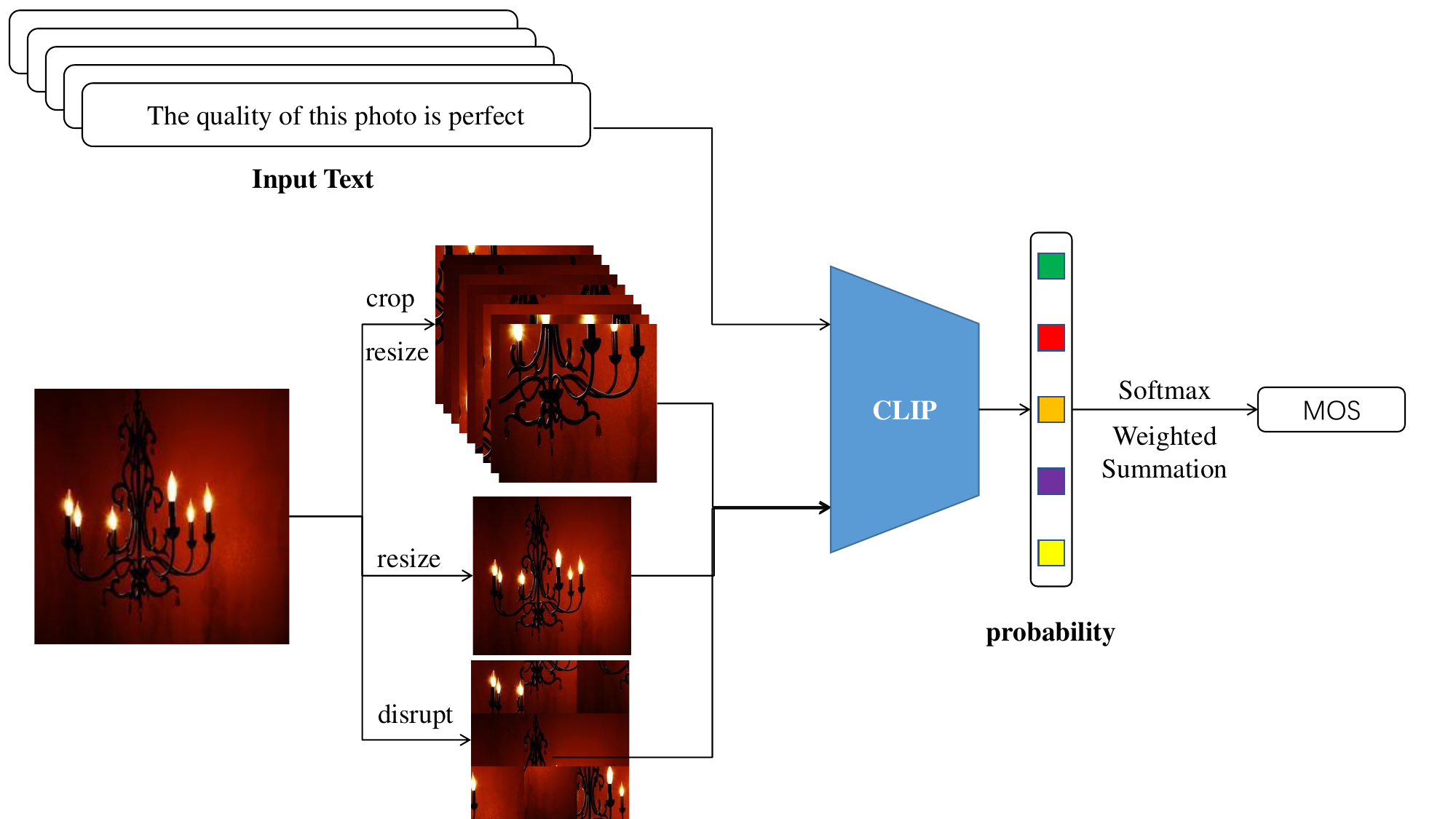}
    \caption{The diagram of the proposed CLIP-IQA*.} 
    \label{fig:model}
    
\end{figure*}

\subsubsection{Implementation Details}

During training and testing, the input data is processed as follows. First, we evenly divide the original image into 9 patches. Secondly, we shuffle the order of the 9 patches and reassemble them into a new image of the original image size. Then, we resize the original image, 9 patches, and the reorganized image to $224 \times 224$. Finally, all the above images are input into the model, and the scores of each image are averaged as the final score.  

During training, the batch size is 3, and the total epochs are set to 80.  We use Smooth-L1 loss as the training loss and CosineAnnealingLR for learning rate decay. In addition, the model with the highest MOS on the validation set is finally selected for testing.

%% file: teams/icl.tex
\subsection{Blind IQA Using Multiple Vision Encoders}
\label{sec:icl}

\emph{Joonhee Lee~$^1$,
Junseo Bang~$^1$,
Se Young Chun~$^{1,2}$} \\
\textit{
$^1$ Department of Electrical and Computer Engineering, \\
$^2$ INMC, Interdisciplinary Program in AI, \\Seoul National University, Republic of Korea\\
Team ICL
}

\vspace{5mm}

In this study, we demonstrate that utilizing various image representations enhances perceptual understanding of images and improves the prediction of Image Quality Assessment (IQA) scores. Four pre-trained encoders are employed as feature extractors, and five Ridge regressors are used to map these features to quality predictions. Specifically, along with the Quality-Aware Encoder and Content-Aware Encoders derived from the existing Re-IQA~\cite{saha2023re}, we added task-specific encoders beneficial to IQA. Using this concept, we calculated the IQA score by linearly summing the outputs from the regressors.

The training dataset consisted solely of the 4K images provided by the challenge. However, using images of large size as input exceeded the computational limits set by the challenge. Therefore, a pre-processing step was implemented to crop the center of images to 320$\times$320 pixels before feeding them into the model. The cropping method varied depending on the encoder requirements. For encoders that required global information (content-aware, scene classification, keypoint detection), the images were first cropped to the largest possible square and then resized to 320$\times$320 pixels. For encoders that required local information (quality-aware), patches of 320$\times$320 pixels were employed without resizing.

During training, the features of the four encoders were regressed using ridge regressors. Five ridge regressors were trained; one regressed the features from all encoders, while the other four regressed combinations of features from three encoders each. During inference, the features from the four pre-trained encoders were passed through the five ridge regressors to yield five scores. Each score was weighted and combined to determine the final score (MOS).

\begin{figure*}[t]
    \centering
    \includegraphics[width=1\linewidth]{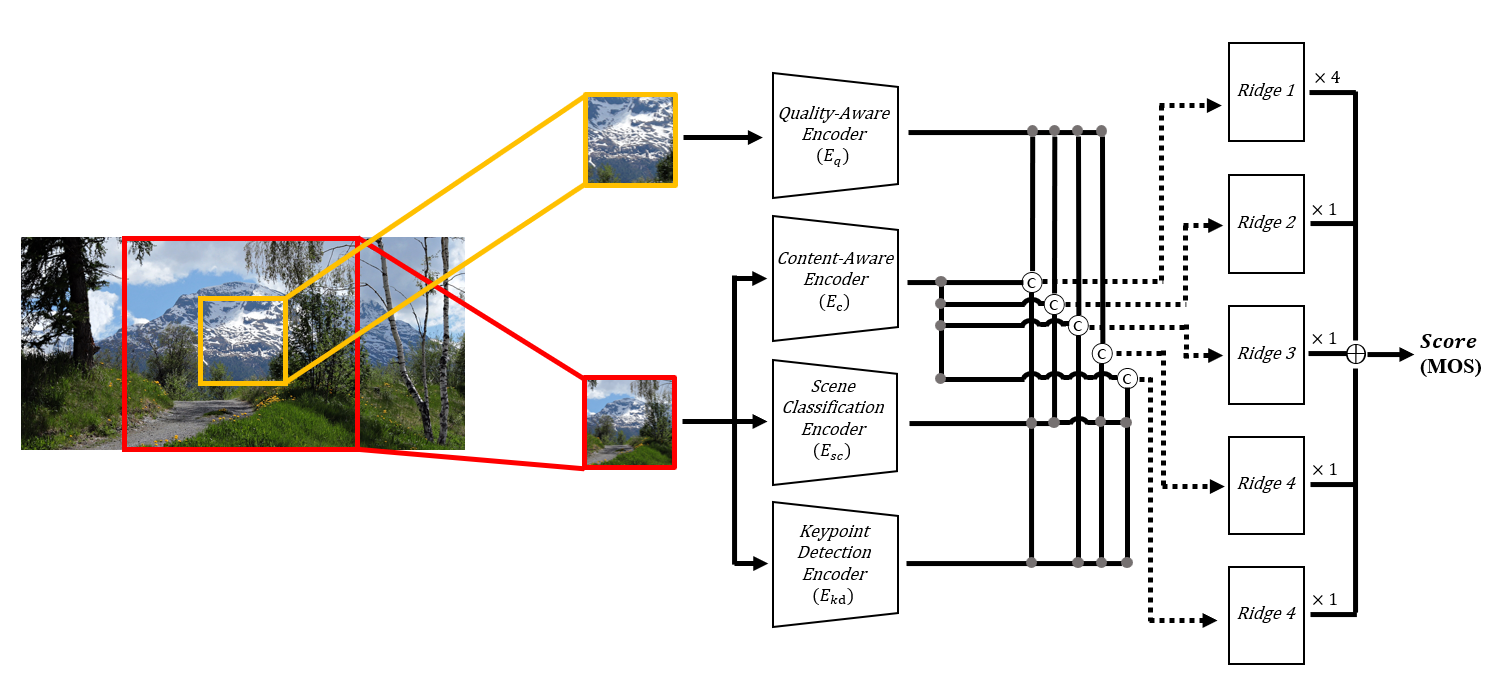}
    \caption{ICL team overall architecture. We use four pre-trained encoders as feature extractors and five ridge regressors to map these features to quality predictions.}
    \label{fig:icl}
\end{figure*}

\subsubsection{Implementation Details}

We optimized the Ridge regression model using the ``GridSearchCV'' function of Scikit-learn~\cite{pedregosa2011scikit}. The hyperparameter alpha was scanned from $10^{-6}$ to $10^{6}$, with 13 equally spaced values on a log scale. We used the entire challenge dataset, dividing the labeled training dataset into a 0.8/0.2 split for training and validation data. With only the ridge regressors being optimized, the training time took approximately 5 to 6 minutes based on NVIDIA A100, the number of parameters is 139.1M, and MACs are 42.09G.